\documentclass[11pt]{article}
\def\bSig\mathbf{\Sigma}

\usepackage[figuresright]{rotating}
\usepackage{longtable}
\usepackage[dvipsnames]{xcolor}
\usepackage{natbib}

\usepackage{booktabs}
\usepackage{graphicx}
\usepackage{rotating}
\usepackage{caption}
\usepackage{subcaption}
\usepackage{amsmath}
\usepackage{graphicx}
\usepackage[colorlinks=true, allcolors=blue]{hyperref}
\usepackage{hyperref}
\usepackage{mathrsfs}
\usepackage{amsfonts}
\usepackage{booktabs} % For \toprule, \midrule, \botrule
\usepackage{threeparttable} % For table footnotes
\usepackage{algorithm}
\usepackage{algorithmicx}
\usepackage{algpseudocode}
\usepackage{stfloats}


\author{
Qiyuan Shi\thanks{Email: qsh@umich.edu} \and
Jian Kang\thanks{Email: jiankang@umich.edu} \and
Yi Li\thanks{Email: yili@umich.edu} \\
Department of Biostatistics, University of Michigan, Ann Arbor, MI, USA
}

\title{Kernel-Based Learning of Chest X-ray Images for Predicting ICU Escalation among COVID-19 Patients}

\begin{document}
\date{}

\maketitle

\begin{abstract}

Kernel methods have been extensively utilized in machine learning for classification and prediction tasks due to their ability to capture complex non-linear data patterns. However, single kernel approaches are inherently limited, as they rely on a single type of kernel function (e.g., Gaussian kernel), which may be insufficient to fully represent the heterogeneity or multifaceted nature of real-world data.
Multiple kernel learning (MKL) addresses these limitations by constructing composite kernels from simpler ones and integrating information from heterogeneous sources. Despite these advances, traditional MKL methods are primarily designed for continuous outcomes. We extend MKL to accommodate the outcome variable belonging to the exponential family, representing a broader variety of data types, and refer to our proposed method as generalized linear models with integrated multiple additive regression with kernels (GLIMARK). Empirically, we demonstrate that GLIMARK can effectively recover or approximate the true data-generating mechanism. We have applied it to a COVID-19 chest X-ray dataset,  predicting binary outcomes of ICU escalation and extracting clinically meaningful features, underscoring the practical utility of this approach in real-world scenarios.
\end{abstract}


%-------------------------------------------
% Paper Body
%-------------------------------------------

%--- Section ---%
\section{Introduction}

 Medical images, such as chest X-rays (CXR) and CT scans, are widely used for diagnosing diseases, evaluating prognoses, and predicting outcomes. Their use has been especially prevalent among COVID-19 patients \citep{benmalek2021comparing, cheng2022covid}. However, they are less frequently explored in the context of predicting escalation to intensive care units (ICU), even though some studies have shown that both CXR scores and baseline CT scans are correlated with ICU admission. During the pandemic, ICUs faced severe resource shortages, making timely and accurate escalation decisions crucial to optimizing care and allocating resources effectively \citep{meille2023covid, arabi2022surging}. CXR has key advantages over CT, being faster, portable, and more accessible, making it a practical tool for real-time decision-making in such high-pressure situations  \citep{jacobi2020portable}. 

Michigan Medicine, a major health provider in the Midwest, maintains multimodal datasets of over five million patients,  including EHR data and CXR imaging data. The CXR imaging dataset contains almost two million images from 178,071 patients admitted to Michigan Medicine patients between 2017 and 2023. With CXR images integrated with the EHR dataset, the data provide access to ICU status and other patient-level information for more comprehensive analysis. A natural question is whether and how we can use the CXR images along with other clinical data, as recorded at the time of patient admission, to predict ICU utilization. The results could potentially inform early interventions and improve patient outcomes in the long term.

Quantifying CXR images can be approached in various ways such as scoring systems and feature-based approaches. Albeit relatively easy, scoring systems lack consistency due to the existence of multiple frameworks, each with its own criteria and interpretation \citep{wasilewski2020covid}. Feature-based approaches, which include radiomics and deep learning techniques, offer a more data-driven alternative. Radiomic features, in particular, are often handcrafted by extracting quantitative, reproducible characteristics from medical images—an approach that is highly desirable and has been successful in many COVID-19 imaging applications \citep{sun2023use}.

One key preparation in modeling radiomics features is proper feature mapping. This is crucial for uncovering complex patterns and capturing non-linear relationships in the data. Traditional single-kernel methods may fail to capture the full data complexity, whereas Multiple Kernel Learning (MKL) \citep{gonen2011multiple} offers a more flexible approach by combining candidate kernels to better represent diverse patterns. Recent advancements in MKL have focused on various aspects, including kernel optimization, computational efficiency, and application to different domains. For example, \citet{liu2019absent} addresses the missing data problem, \citet{shen2021distributed} enables online learning, and \citet{han2018matrix} controls model complexity.

However, most MKL extensions primarily focus on squared loss and machine learning methods like support vector machines (SVMs), rather than probabilistic models, limiting its applicability within the broader statistical toolkit designed for diverse data types. Another challenge in MKL is dimension reduction (DR). Common methods in DR have been integrated with MKL such as MKL-DR \citep{lin2010multiple} but they still largely depend on projection or embedding, which compromises interpretability. Interpreting original features has received some attention \citep{briscik2023improvement}. 
To address interpretability, we focus on the Multiple Additive Regression Kernels (MARK) \citep{bennett2002mark}, which was based on column generation (CG) \citep{demiriz2002linear}. One key advantage is that, instead of searching for an optimal kernel, it focuses on feature-kernel combinations (i.e., columns), preserving the interpretability of the original features. Additionally, the selection process primarily involves matrix multiplication, a workload that modern GPUs, with their rapid advancements, handle efficiently.

Nonetheless, the original MARK method has notable limitations. First, it implicitly assumes Gaussian-type outcomes by relying solely on squared loss functions, raising uncertainty about its applicability to binary outcomes, as is the case in our study. Second, although the method scales nearly linearly with sample size, it struggles with a large number of feature-kernel combinations. This limitation is significant because, unlike datasets typically used in MKL studies \citep{breast_cancer_wisconsin_(diagnostic)_17, harrison1978hedonic}, radiomics often involves regions of interest (ROIs)—specific anatomical or pathological areas within an image, such as lung lobes, tumors, or fibrotic regions—where different regions may require distinct representations. This can result in a substantial computational burden. Given a sample size of $N$, $P$ partitions of features, and $Q$ types of kernels, the full kernel matrix $\boldsymbol{K}$ consists of $NPQ$ columns. To illustrate the scale, the Boston Housing dataset has $D = 13$ features, yielding at most $2^{13} = 8192$ possible subsets. However, most of these combinations do not reflect meaningful groupings, limiting their utility for structured modeling. In contrast, our radiomics dataset has $D = 2046$, resulting in an astronomical number of possible partitions ($2^{2046}$), many of which correspond to anatomically or clinically relevant groupings, making them more appropriate for kernel assignment.

To address these limitations for analyzing CXR images as presented in our dataset, we propose a modified MARK, termed  Generalized Linear Models with Integrated
Multiple Additive Regression with Kernels (GLIMARK),  that offers improvements in the following aspects:
1) we introduce a likelihood-based loss function within the generalized linear model (GLM) framework, accommodating a range of outcomes, including binary, Poisson, and Gaussian; 2)  we propose a hierarchical framework for modeling CXR radiomics features. The proposed GLIMARK explores feature patterns at multiple scales, balancing model complexity and pattern learning capacity. Additionally, the modification with the ${\mathcal{H}_k}$-norm can give a more rigorous explanation aligned with the representer theorem. Our analysis focuses on the binary outcome of ICU escalation. We have applied our method to analyze the data 
and found that the inclusion of multiple kernels can uncover additional important radiomics features. Specifically, the Neighboring Gray Tone Difference Matrix (NGTDM) and Shape2D features are identified as top-ranked feature classes, which are not detected by either traditional regression models or random forest (RF) models. In addition, the `representer' patients highlight key visual features for further radiological analysis, serving as benchmarks for predicting outcomes among future patients.

In Section 2, we review the related work in MKL. In Section 3, we introduce the proposed GLIMARK. In Section 4, we perform simulation studies to show that if the outcome is generated by some true functional forms, GLIMARK can recover or approximate the truth.  In Section 5, we apply the method to real CXR data from Michigan Medicine hospital system. In Section 6, we discuss the strengths and limitations of our approach, as well as its potential with lateral view CXRs and Vision Transformers.

\vspace{-0.5cm}

\section{Related Work and Motivation}
\subsection{Representer results with non-Gaussian Outcome Data}
  
Given a training sample of \( N \) independent individuals, for $i = 1,\ldots, N$,  let \( \boldsymbol{z}_i \) represent the predictor (e.g., the CXR image data) and \( y_i \) denote the outcome (e.g., whether the hospitalized patient escalates into ICU).  It follows that the conditional distribution of \( y_i \) given \( \boldsymbol{z}_i \) can be specified by a functional parameter \( f(\cdot) \) induced from a reproducing kernel Hilbert space (RKHS) \( \mathcal{H}_k \), 
which admits the form: 
\[
f(\cdot) = \sum _{j=1}^{N}\alpha_{j} k(\boldsymbol{z}_j, \cdot) + b,
\] 
where \( k (\cdot, \cdot)\) is the reproducing kernel associated with \( \mathcal{H}_k \). 
The representation follows from the \emph{representer theorem} \citep{wahba2019representer}, which states that for many regularized learning problems in RKHS, including
\eqref{objective} in our case, the solution can be expressed as a finite linear combination of kernel evaluations at the training points. Such representations are widely used in applications, including kernel ridge regression (KRR) \citep{lin2020distributed}, kernel Cox models \citep{rong2024kernel}, and deep learning \citep{unser2019representer}.

\subsection{Kernel Construction} \label{sec:kernels}
One main goal of MKL is to construct a positive definite kernel from pre-defined kernels. The construction could be linear, non-linear or even data dependent \citep{gonen2011multiple}. For example, the sum of 
% if $k_1(\cdot, \cdot)$ and $k_2(\cdot, \cdot)$ 
two valid kernels 
is also a valid kernel,
%is $ k_3(\cdot, \cdot)= k_1(\cdot, \cdot) + k_2(\cdot, \cdot).$ 
motivating us to use  conic combinations  construct new kernel functions. \citet{lanckriet2004learning} have successfully improved SVM performances using combinations of linear, polynomial and radial basis function (RBF) kernels with the decision function: 
\(
f(\boldsymbol{z}) = \sum_{j=1}^{N} \alpha_{j} k(\boldsymbol{z}_{j}, \boldsymbol{z}) + b,
\) 

where $
k(\boldsymbol{z}_{j}, \boldsymbol{z}) = \sum_{q=1}^{Q} \mu_{q} k_{q}(\boldsymbol{z}_{j}, \boldsymbol{z}).$ %only use display mode is necessary.
Here $\boldsymbol{z}_{j} \in \mathbb{R}^D$ are the training points, $\boldsymbol{z} \in \mathbb{R}^D$ is any arbitrary point to make prediction on, $Q$ is the number of kernels and $\mu_q \ge 0, q=1, \ldots, Q, $ are the weights of candidate kernels.  

This construction of combined kernels can effectively handle various  data sources. Suppose the predictors consist of \( P \) components or partitions, each capturing a distinct aspect of the data. For instance, in CXR radiomics, features are extracted from two ROIs, each processed with eleven filters and grouped into seven feature classes. This yields a total of 2046 features, which naturally form 154 partitions based on ROI, filter, and feature class combinations:
\(
P = 2 \times 11 \times 7 = 154
\).

By learning the kernel weights \( \mu_p \) from data, we can construct a flexible, data-driven kernel representation:  
\(
k(\boldsymbol{z}_{j}, \boldsymbol{z}) = \sum_{p=1}^{P} \mu_{p} k_{[p]}(\boldsymbol{z}_{j}^{p}, \boldsymbol{z}^{p}),
\)  
where \( \boldsymbol{z} = \{\boldsymbol{z}^{p}\}_{p=1}^{P} \) and \( \boldsymbol{z}_{j} = \{\boldsymbol{z}_{j}^{p}\}_{p=1}^{P} \).  
When $P$ is large, instead of forming \( P \) new kernels (which may be impractical), a more effective approach is to use a common set of \( Q \) kernels across all partitions:  
%\begin{equation} \label{eqn:k}
\(k(\boldsymbol{z}_{j}, \boldsymbol{z}) = \sum_{q=1}^{Q} \sum_{p=1}^{P} \mu_{p,q} k_{q}(\boldsymbol{z}_{j}^{p}, \boldsymbol{z}^{p}).
\)
This formulation, which allows adaptation of kernels to heterogeneous data sources, aligns with multi-view learning \citep{yan2021deep}. In image analysis, partitions may correspond to different ROIs (e.g., lung vs peripheral area), different aspects such as texture and shape, different imaging perspectives (e.g., posteroanterior (PA) vs. lateral (LAT) CXR), or even distinct imaging modalities (e.g., CXR vs. CT). Inspired by multi-view learning paradigm, we refer to a kernel-partition combination as a `view'

(Figure~{\ref{fig:concept}}).

\section{  Generalized Linear Models with Integrated Multiple Additive Regression with Kernels (GLIMARK)}
\subsection{Method}

%\qsol{Introducing GLM setting}
To accommodate non-Gaussian outcome data (e.g., binary ICU use) as encountered in our data example, we extend MKL to a generalized linear model (GLM) setting \citep{bennett2002mark,zhu2005kernel}. Assume the outcome \( y_i \) follows a distribution from the exponential family, whose probability density function is given by:
\begin{equation} \label{glm-pdf}
p(y|\theta) = \exp\left(y\theta - b_{glm}(\theta) + c_{glm}(y) \right),
\end{equation}
where for simplicity we omit the dispersion parameter as in \cite{fei2021estimation}, because our focus is on the mean parameters.
Apart from a constant that depends only on the observed data, the log-likelihood for a single observation is given by:
\(
\ell(\theta_{i} | y_{i}) = y_{i} \theta_{i} - b_{glm}(\theta_{i}), \quad i = 1, \dots, N,
\)
where \( b_{glm}(\cdot) \) is a function linking the mean of \( y_i \) to \( \theta_i \). Under the canonical link function for GLMs, the decision function \( f \) is linked to the expectation of the outcome via:
\[
\theta_i = f_i = g_{glm}(\mu_i), \quad \text{where } f_i = f(\boldsymbol{z_i}) \text{ and } \mu_i = \mathbb{E}[Y_i | \boldsymbol{z_i}].
\]
%\qsol{Introducing GLM with regularized empirical risk function }
Hereafter, we use \( \theta_i \) and \( f_i \) interchangeably.
Consequently, we propose to use the following  regularized empirical risk function:
\begin{equation} \label{objective}
H = -\frac{1}{N} \sum_{i=1}^{N} \ell(f_i | y_i) + \lambda P(\| f \|_{\mathcal{H}_k})
= -\frac{1}{N} \sum_{i=1}^{N} \left[ y_i f_i - b_{glm}(f_i) \right] + \lambda P(\| f \|_{\mathcal{H}_k}),
\end{equation}
where \( \lambda > 0 \) is a regularization parameter controlling the trade-off between model fit and complexity, \( P(\cdot) \) is a monotonically increasing function, and \( H \) is the objective function to be minimized.
Here, the RKHS norm of \( f \) is denoted as \( \| f \|_{\mathcal{H}_k} \).  
%\qsol{Plugging MKL into GLM with regularized empirical risk function }
With a training sample size of \( N \), a natural extension of MKL is to form a decision function as:
\begin{equation} \label{junk}
f(\boldsymbol{z}) = \sum_{j=1}^{N} \alpha_{j} \sum_{q=1}^{Q} \sum_{p=1}^{P} \mu_{p,q} k_{q}(\boldsymbol{z}_{j}^{p}, \boldsymbol{z}^{p}) + b.
\end{equation} 
%Many methods focus on estimating \( \boldsymbol{\mu} \) along with \( \boldsymbol{\alpha} \), 
But there are several issues.
First, to address the identifiability problem, 
additional constraints are necessary, such as 
\(
\sum_{p=1}^{P} \sum_{q=1}^{Q} \mu_{p,q} = 1,
\)
  which, however, may complicate optimization, particularly when the RKHS norm \( \| f \|_{\mathcal{H}_k} \) depends on both \( \boldsymbol{\alpha} \) and \( \boldsymbol{\mu} \).  Moreover, as Kernel methods are inherently similarity-based, meaning that making a prediction for a new point \( \boldsymbol{z} \) requires computing similarities between \( \boldsymbol{z} \) and all training points \( \boldsymbol{z}_j \) for \( j = 1, \dots, N \) across the \( PQ \) views. The computation is expensive, particularly when multiple kernels are involved.  Multiple Additive Regression with Kernels (MARK) \citep{bennett2002mark} addresses these issues by removing any constraints on the coefficients and re-parameterizing $\boldsymbol{\alpha}$ to effectively reduce MKL problem to a standard regression model:
  \begin{equation}
\label{junk2}
f(\boldsymbol{z}) = \sum_{j=1}^{N} \sum_{q=1}^{Q} \sum_{p=1}^{P} \alpha_{j,p,q} k_{q}(\boldsymbol{z}_{j}^{p}, \boldsymbol{z}^{p}) + b.
\end{equation}
To deal with overparameterization, MARK only selects a limited set of $\alpha_{j,p,q}$ or their corresponding patient-view combinations. %See section 3 for details.  

As few MARK-based methods are available for non-Gaussian outcome data,  we 
extend it to accommodate outcome data that belong to the exponential family \eqref{glm-pdf}.
Following the idea of MARK and column generation (CG) \citep{lubbecke2005selected}, we design an algorithm that selects patient-view combinations represented by the columns of a design matrix. Specifically, we  construct a design matrix corresponding to  Model (\ref{junk2}) that concatenates kernel matrices across all \( P \) partitions and \( Q \) kernels:

\begin{equation}
    \boldsymbol{K} =
    \begin{bmatrix}
\boldsymbol{K}_{1,1}  & \dots & \boldsymbol{K}_{1,Q} &
        \boldsymbol{K}_{2,1} & \dots & \boldsymbol{K}_{2,Q} &
        \dots &
        \boldsymbol{K}_{P,1} & \dots & \boldsymbol{K}_{P,Q}
    \end{bmatrix},
    \label{eqn:stack}
\end{equation}
where  each block \( \boldsymbol{K}_{p,q} \) is  an $N \times N$ matrix and    each entry of   \( \boldsymbol{K}_{p,q} \) is given by  $k_q(\boldsymbol{z_i}^{p}, \boldsymbol{z_j}^{p}), i,j =1, \ldots, N$, where \( k_q(\cdot, \cdot) \) is as defined in Section \ref{sec:kernels}.  Accordingly, each column in (\ref{eqn:stack}) is a patient-view combination.
Then we express Model   \eqref{junk2}  (for any $\boldsymbol{z}_{i} \in \mathcal{X}$) as:
\begin{equation}
\label{eqn:eq2}
 f_{i}=f(\boldsymbol{z}_{i})=\sum_{j=1}^{J}  \alpha_{j}[\boldsymbol{K}]_{i,j} + b,
\end{equation}
where $J=NPQ$ and $[\boldsymbol{K}]_{i,j}$ is the $(i,j)$-th entry of $\boldsymbol{K}$.

We refer to the algorithm  as Generalized Linear Models with Integrated Multiple Additive Regression with Kernels (GLIMARK), which is detailed below. GLIMARK iteratively builds a sparse predictive model by repeating a selecting and updating process at each iteration: it selects one component of $\boldsymbol{\alpha} \in \mathbb{R}^{J}$ (or equivalently the corresponding column of $\boldsymbol{K}$),  followed by updating the estimates of all the selected components to reduce Eq.~(\ref{objective}) until a pre-specified number of steps or the improvement is smaller than some threshold,  e.g.,  $tol=10^{-6}$. Here, $\boldsymbol{\alpha}^{(t)} \in \mathbb{R}^{J}$ and $b^{(t)}$ denote the estimates of  $\boldsymbol{\alpha}$ and $b$ at iteration $t$.  This process is essentially a forward selection method, as it begins with the null model  \(\boldsymbol{\alpha}^{(0)} = \boldsymbol{0}\), and  at each iteration, a new column is sequentially added based on the impact of \(\alpha_j\) on Eq.~(\ref{objective}) from the remaining candidate variables:
\begin{equation}
\begin{split}
\frac{\partial H(\boldsymbol{\alpha}, b)}{\partial \alpha_{j}}
%\Big|%_{(\boldsymbol{\alpha} = \boldsymbol{\alpha}^{(t)},  b= b^{(t)})} 
=&  -\frac{1}{N}\sum_{i=1}^{N} \left[ y_{i} - \frac{\partial b_{glm}(f_{i}(\boldsymbol{\alpha}, b))}{\partial f_{i}} %\Big|
%_{(\boldsymbol{\alpha} = \boldsymbol{\alpha}^{(t)}, b= b^{(t)})} 
\right] [\boldsymbol{K}]_{i,j} \\
& + \lambda \frac{\partial P(\|f(\boldsymbol{\alpha}, b)\|_{\mathcal{H}_k})}{\partial \alpha_{j}} 
%\Big|%_{(\boldsymbol{\alpha} = \boldsymbol{\alpha}^{(t)}, b= b^{(t)})}.
\end{split}
\end{equation}
with all the partial derivatives evaluated
at $(\boldsymbol{\alpha}, b) = (\boldsymbol{\alpha}^{(t)}, b^{(t)})$.

The selection results are denoted by the index set $S^{(t)}$ where \(S^{(t)} = (j^{*(1)},j^{*(2)} \ldots, j^{*(t)})\) and $S^{(0)}=\emptyset$. At each step we select one unselected component of $\boldsymbol{\alpha}$ indexed by \(j^{*(t+1)}\) and update \(S^{(t+1)}\), where
\begin{equation}
\label{selection}
    j^{*(t+1)} = \arg\max_{j : j \notin S^{(t)}} 
    \left| 
         \frac{\partial  H(\boldsymbol{\alpha}, b)}{\partial \alpha_j} 
        \right|
\end{equation}
and the partial derivative is evaluated as  $(\boldsymbol{\alpha}, b) = (\boldsymbol{\alpha}^{(t)}, b^{(t)})$. 
In essence, the absolute value of the partial derivative serves as the criterion for forward selection.
% Meanwhile, we select column \(j^{*(t+1)}\) of $\mathbf{K}$ into the working kernel matrix $\boldsymbol{K}^{(t)}$.

The estimation is implemented by gradient descent based methods such as Adam \citep{kingma2014adam} that minimize Eq.~(\ref{objective}) with:
\begin{equation}
\label{fshort}
f_{i}=\sum_{j \in S^{(t+1)}} \alpha_{j}[\boldsymbol{K}]_{i, j} + b.
\end{equation}
Thus, \(\alpha_j^{(t+1)}\) and $b^{(t+1)}$ are updated for all \( j \in S^{(t+1)}\) using Adam, and \(\alpha_j^{(t+1)} = 0, \quad \forall j \notin S^{(t+1)}\).

\begin{algorithm}  
\caption{GLIMARK} 
\begin{algorithmic}[1]
    \State Let $b^{(0)} = g_{glm}(p_{+})$, $p_{+}= \frac{1}{N}\sum_{i=1}^{N}y_{i}$, $tol=10^{-6}$
    \State Let $S^{(0)} := \emptyset$
    \State Let $\boldsymbol{\alpha}^{(0)} = \boldsymbol{0}$
    \For{$t := 0$ to $T_{max}$}
        % \State  Select kernel column $ [\boldsymbol{K}]_{\cdot, j^{*(t)}}$  in $\boldsymbol{K} \backslash \boldsymbol{K}^{(t)}$ by maximizing $|\frac{\partial H^{(t)}}{\partial \alpha_{j}}|$ in Eq.~(\ref{eqn:eq1})
        \State Select $j^{*(t+1)}$ according to (\ref{selection})
        % \If{$\lvert  \frac{\partial H}{\partial \alpha_{j^{*(t+1)}}}\Big|_{(\boldsymbol{\alpha}=\boldsymbol{\alpha}^{(t)},b=b^{(t)})}\rvert  \leq tol$}
        \If{$\left| \left. \frac{\partial H}{\partial \alpha_{j^{*(t+1)}}} \right|_{(\boldsymbol{\alpha}, b) = (\boldsymbol{\alpha}^{(t)},\; b^{(t)})} \right| \leq tol$}

            \State \textbf{return} $\boldsymbol{\alpha}^{(t)}, b^{(t)}$ as the final estimates, i.e.,  $\boldsymbol{\hat{\alpha}}$ and $\hat{b}$
        \EndIf
        % \State $\boldsymbol{K}^{t+1} = [\boldsymbol{K}^{(t)} [\boldsymbol{K}]_{\cdot, j^{*(t)}}]$
        % \State $\boldsymbol{K}^{(t+1)} = [\boldsymbol{K}^{(t)} \; [\boldsymbol{K}]_{\cdot, j^{*(t)}}]$
        \State Update $S^{(t+1)}$ as  $S^{(t+1)} = S^{(t)} \cup \{j^{*(t+1)}\}$
        \State Minimize Eq.~(\ref{objective}) with (\ref{fshort}) using Adam, give updates for \(\alpha_j^{(t+1)},\quad \forall j \in S^{(t+1)}\) and $b^{(t+1)}$.
        Let \(\alpha_j^{(t+1)} = 0, \quad \forall j \notin S^{(t+1)}\)
    \EndFor
    \State \textbf{return} $\boldsymbol{\alpha}^{(t+1)}, b^{(t+1)}$ as the final estimates, i.e., $\boldsymbol{\hat{\alpha}}$ and $\hat{b}$
\end{algorithmic} 
\end{algorithm}
   The proposed GLIMARK is summarized by Algorithm 1 and is  detailed in Web Appendix A. The algorithm supports a variety of response data types. As an example, with  $P(\cdot)$  chosen  the square function,  the objective function $H$ in the algorithm can be specified for the binomial and Poisson distributions as:
\[
H(\boldsymbol{\alpha}, b) =
    \begin{cases} 
        -\frac{1}{N} \sum_{i=1}^{N} \left[y_{i} f_{i} - \log(1+\exp(f_{i})) \right] + \lambda\|f\|_{\mathcal{H}_k}^{2}, & \text{(Binomial)} \\
        -\frac{1}{N} \sum_{i=1}^{N} \left[y_{i} f_{i} - \exp(f_{i}) \right] + \lambda\|f\|_{\mathcal{H}_k}^{2}, & \text{(Poisson)}
    \end{cases}
\]
with the gradients  given by:
\[
    \frac{\partial H(\boldsymbol{\alpha}, b)}{\partial \alpha_{j}} =
    \begin{cases} 
        -\frac{1}{N} \sum_{i=1}^{N} \left[y_{i} - \frac{\exp(f_{i})}{1+\exp(f_{i})} \right] [\boldsymbol{K}]_{i,j} + \lambda \frac{\partial \|f\|_{\mathcal{H}_k}^{2}}{\partial \alpha_{j}}, & \text{(Binomial)} \\
        -\frac{1}{N} \sum_{i=1}^{N} \left[y_{i} - \exp(f_{i}) \right] [\boldsymbol{K}]_{i,j} + \lambda \frac{\partial \|f\|_{\mathcal{H}_k}^{2}}{\partial \alpha_{j}}. & \text{(Poisson)}
    \end{cases}
\]

\subsection{Hierarchical Structures, Interpretation and Implementation}

\begin{figure}
    \centering
    \begin{subfigure}{\textwidth}
        \centering
        \includegraphics[width=\linewidth]{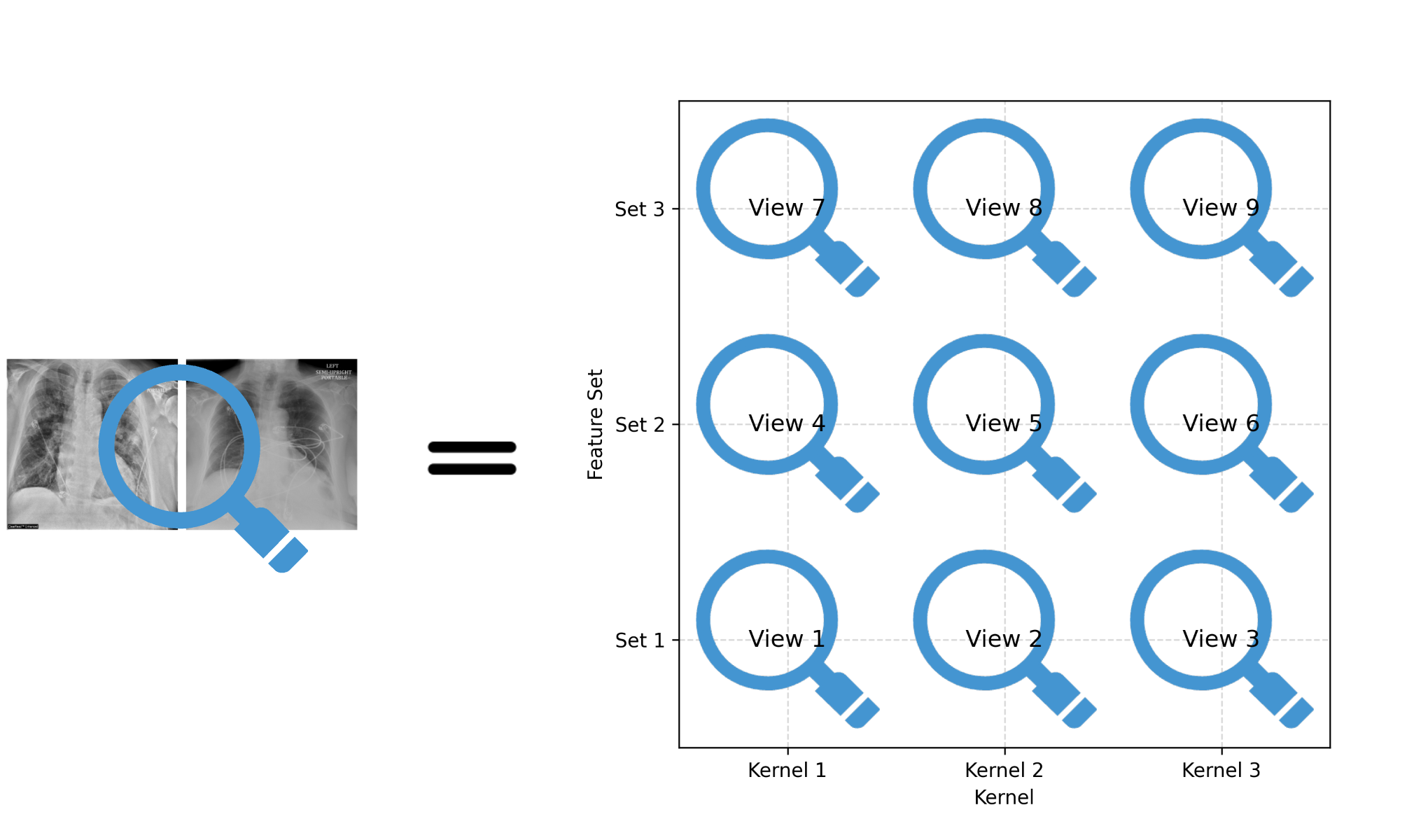}
        \caption{A multi-view comparison of two images.}
        \label{(a)}
    \end{subfigure}\\[1em]
    \begin{subfigure}{\textwidth}
        \centering
        \includegraphics[width=\linewidth]{figures/fig_concept2.png}
        \caption{Hierarchical model structures. }
        \label{(b)}
    \end{subfigure}
    \caption{}
    \label{fig:concept}
\end{figure}

SVM methods interpret the decision function as a weighted sum of similarity measures between inputs and support vectors, contrasting with traditional regression, which focuses on explaining variance. The representer theorem bridges these approaches by showing that the optimal decision function, though defined over infinitely many variables, can be expressed as a linear combination of "representers." Inspired by this, MKL measures similarity across multiple feature spaces, with each feature space defined by a view—combinations of kernels and feature subsets that capture specific perspectives (Figure~{\ref{fig:concept}}). In GLIMARK, the decision function becomes a weighted sum of similarities between new observations and selected patient-view combinations, corresponding to columns of the concatenated matrix $\boldsymbol{K}$. Views exhibit a hierarchical structure, applying kernels to single attributes or groups, akin to weak and strong learners in the MARK framework.

Since kernels operate at different scales and regularization is achieved via \(\|\boldsymbol{\alpha}\|_{2}\) or \(\|f\|_{\mathcal{H}_k}\), a two-level GLIMARK resembles hybrid group sparse lasso methods, such as in \citep{friedman2010note}:
\(
\min_{\boldsymbol{\alpha}} -\frac{1}{N}\sum_{i=1}^{N}\ell(f_{i}|y_{i}) + \lambda_1 \sum_{\ell=1}^L \|\boldsymbol{\alpha}_{\ell}\|_2 + \lambda_2 \|\boldsymbol{\alpha}\|_1.
\)
This framework induces competition between features at different scales, enabling the model to select appropriate scales automatically. Unlike traditional methods, a two-level GLIMARK uses each variable twice: once at the group level and once individually, potentially reducing coefficient magnitudes for some scales.

In simple cases like kernel SVM classification, the relationship between \(\boldsymbol{\alpha}\) (dual) and feature weights \(\boldsymbol{w}\) (primal) is given by \citep{bishop2006pattern}:
\(
\boldsymbol{w} = \sum_{i=1}^{N} \alpha_{i} y_{i} \phi(\boldsymbol{z}_{i}),
\)
where \( f(\boldsymbol{z}) = \boldsymbol{w} \cdot \phi(\boldsymbol{z}) + b \) in the primal problem. The feature mapping \(\phi(\cdot)\) can be obtained theoretically or empirically \citep{scholkopf1999input, minh2006mercer} from kernels. For GLIMARK, these relationships are less direct. To assess variable detection performance, we propose flexibly summing \(| \hat{\alpha}_{j}|\) by groups as $I_{\text{Group}}$. For instance, $I_{\text{First-Order\_L\_RBF}}$ is marginal over all individuals and shows overall importance of views from First-Order features in the lung area and RBF kernels, whereas $I_{\text{First-Order}}$ is overall importance of First-Order features. Alternatively, $I_{\text{PatientID}}$ sums over views for each individual and identifies important representers. A high value indicates either broad comparisons across feature spaces or a significant weight in one space, highlighting key training points for further analysis.

 For classification performance, GLIMARK predicts binary outcome probabilities, allowing ROC curve generation. We use Youden's index \citep{fluss2005estimation}, defined as sensitivity plus specificity minus one, to determine the optimal threshold for classification, though a 0.5 cutoff can also be applied. In cross-validation with hierarchical data, two hyperparameters are tuned: \(\lambda\), which regularizes model complexity, and \(T\), which determines the number of terms for optimal performance.  Accuracy based on Youden's index, the 0.5 cutoff, AUC, or F-score can guide fine-tuning of these parameters.

% \begin{figure*}

% \end{figure*}

\section{Simulation Study}
The simulation  serves two major aims. Our first aim  
is to determine its ability to detect potential linear and nonlinear relationships between the outcome and predictors, as well as any possible interactions. Our second aim is to evaluate the model’s classification performance relative to other common methods, such as SVMs and random forests, and also to explore the optimal combination of the parameters $\lambda$ and $T$, both of which control the complexity of the model.   

 For our first aim, we follow a setup similar to \citet{friedman2010note}, with 10 groups of variables (G1 to G10), each containing 10 variables generated from a standard Gaussian distribution. There are no correlations within or between groups. The number of active variables in G1 through G6 is (10, 8, 6, 4, 2, 1), while G7 to G10 are irrelevant. For observation \( \boldsymbol{z}_i \), \( \boldsymbol{z}_{i,g} \) represents the active variables in group \( g \), and \( \boldsymbol{\beta_g} \) denotes their corresponding coefficients, randomly chosen as \( \pm 1 \). The scenarios (Table~\ref{tbl:tbl1}) include: Models A, B, and C, which involve linear or non-linear relationships, and Models D and E, which examine interaction effects.

\begin{figure} % Use [H] for exact placement
\centerline{%
\includegraphics[width=\linewidth]{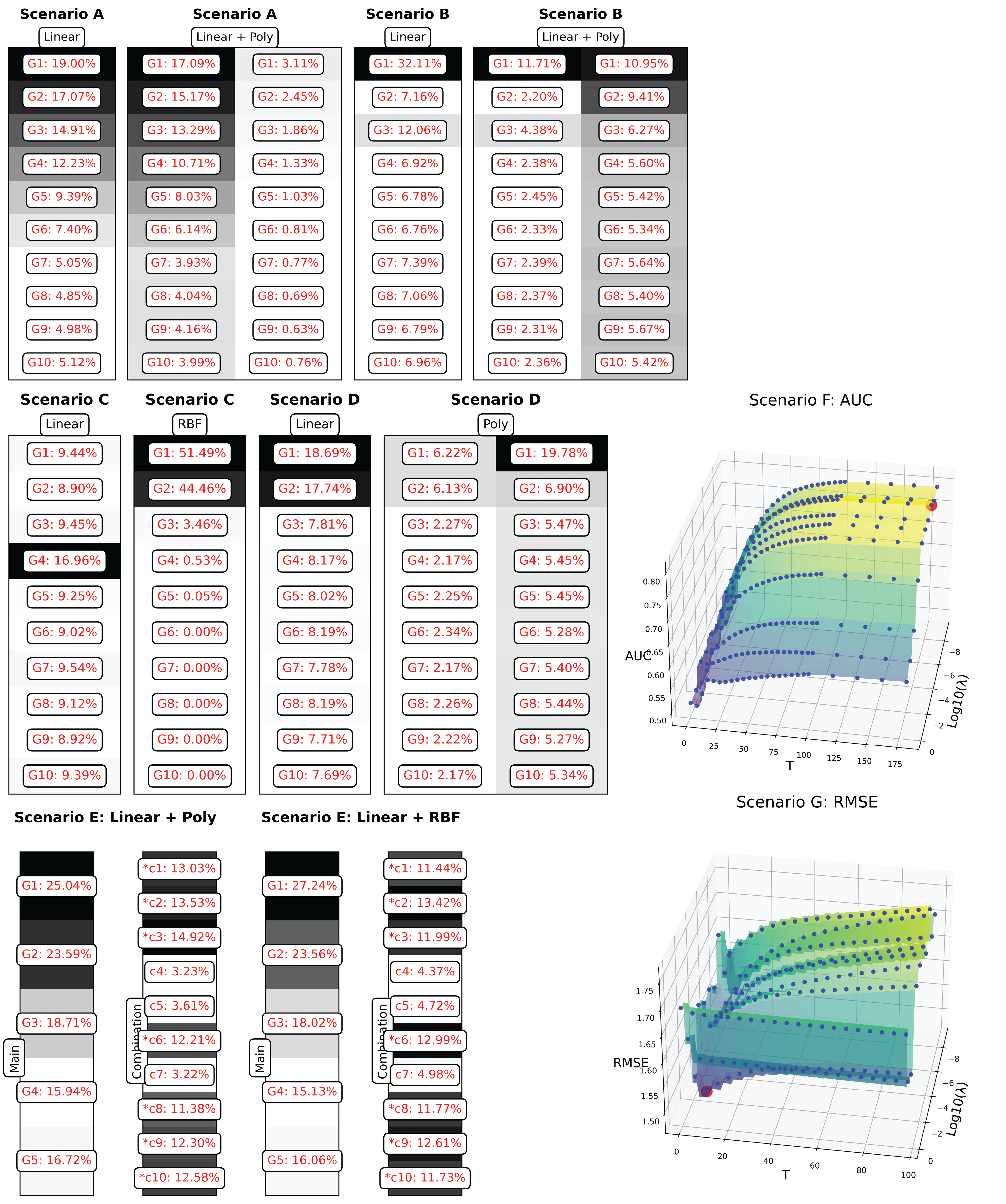}}
  \caption{Simulation results. Red text represents the contribution percentage of each view. \textbf{G\#} denotes a group, while \textbf{C\#} refers to a possible combination. Asterisks indicate the truly active combinations in the true model. Scenarios A–E represent the feature detection performance of GLIMARK, while scenarios F–G represent its overall performance in response to the two hyperparameters $\lambda$ and $T$.}
  \label{fig_sim}
\end{figure}

For each scenario, we conduct 100 independent simulations with a sample size of 400. We fix  \( \lambda = 0.001 \) and \( T = 400 \) when implementing GLIMARK. Results are shown in Figure~\ref{fig_sim}.

\begin{table}
\caption{Simulation settings. Transformations include element-wise cubic (\( \boldsymbol{z}_{i,g}^3 \)), square (\( \boldsymbol{z}_{i,g}^2 \)), exponential (\( \boldsymbol{z}_{i,g}^{exp} \)), and individual variable terms (\( \boldsymbol{z}_{i,g1} \) for the first variable in group 1). Architecture indicates whether kernels are applied to groups (G), individuals (I), or both (G + I).}
\label{tbl:tbl1}
\centering
\begin{tabular}{@{}ccc@{}}
    \toprule
    \textbf{Scenario} & \textbf{Nonlinear Relationship} & \textbf{Architecture} \\ \midrule
    A & \( f(\boldsymbol{z}_{i}) = \sum_{g=1}^{6} \langle \boldsymbol{z}_{i,g}, \boldsymbol{\beta}_g \rangle + 1 \) & G \\ 
    B & \( f(\boldsymbol{z}_{i}) = \langle \boldsymbol{z}_{i,1}^{3}, \boldsymbol{\beta}_1 \rangle + \langle \boldsymbol{z}_{i,2}^{2}, \boldsymbol{\beta}_2 \rangle + \langle \boldsymbol{z}_{i,1}, \boldsymbol{\beta}_1 \rangle + 1 \) & G \\ 
    C & \( f(\boldsymbol{z}_{i}) = 3\sum_{g=1}^{6} \langle \boldsymbol{z}_{i,g}^{exp}, \boldsymbol{\beta}_g \rangle + \langle \boldsymbol{z}_{i,4}, \boldsymbol{\beta}_4 \rangle + 1 \) & G \\ 
    D & \( f(\boldsymbol{z}_{i}) = \langle \boldsymbol{z}_{i,1}, \boldsymbol{\beta}_1 \rangle + \langle \boldsymbol{z}_{i,2}, \boldsymbol{\beta}_2 \rangle + \sum_{j < k} \gamma_{jk} z_{i,1j} z_{i,1k} + 1 \) & G \\ 
    E & \( f(\boldsymbol{z}_{i}) = \langle \boldsymbol{z}_{i,1}, \boldsymbol{\beta}_1 \rangle + \sum_{j < k} \gamma_{jk} z_{i,1j} z_{i,1k} + 1 \) & G + I \\ \midrule

    F & 
    \( \begin{aligned}
    f(\boldsymbol{z}_{i}) &= 3\sin\left( 2\sum_{g=1}^{10} |z_{i,g1}| + 1 \right) + \exp\left( 0.0001 \sum_{g=1}^{10} z_{i,g2} - 2 \right) \\
    &\quad + 100\langle \boldsymbol{z}_{i,3}^{2}, \boldsymbol{\beta}_3 \rangle + \langle \boldsymbol{z}_{i,4}, \boldsymbol{\beta}_4 \rangle + 1
    \end{aligned} \) 
    & G + I \\

    G & 
    \( \begin{aligned}
    f(\boldsymbol{z}_{i}) &= \frac{1}{20} \big[ 10\sin\left( 50\sum_{g=1}^{10} |z_{i,g1}| + 1 \right) + 25\exp\left( 0.1 \sum_{g=1}^{10} z_{i,g2} - 2 \right) \\
    &\quad + 3\langle \boldsymbol{z}_{i,3}^{2}, \boldsymbol{\beta}_3 \rangle + 5\langle \boldsymbol{z}_{i,4}, \boldsymbol{\beta}_4 \rangle + 1 \big]
    \end{aligned} \)
    & G + I \\
    
    \bottomrule
\end{tabular}
\end{table}

% use * for 2col, but NOT for Referee!!!

\textbf{Scenario A (Linear Model):} A linear GLIMARK correctly identifies active groups, with $I_{\text{Group}}$ contributions proportional to the number of active variables (G1-G6: 19\% to 7\%). Minor noise appears in non-active groups (5\%). Adding an RBF kernel does not change detection and only introduces noise (e.g., 3\% in G1-RBF).

\textbf{Scenario B (Cubic and Square Terms):} A linear GLIMARK fails to capture G2, estimating its $I_{\text{Group}}$ at 7\%, similar to noise. Adding a degree-2 polynomial kernel improves detection, increasing G2’s contribution to 9\%, which suggests that the polynomial kernel helps capture relevant nonlinear structure.

\textbf{Scenario C (Linear and Exponential Terms):} The linear GLIMARK detects only the linear effects of G4 but fails to capture exponential contributions from G1-G6. It also introduces noise of 9\%. The additional RBF kernel successfully identifies exponential effects in the first two groups but not the linear effect in G4. This highlights the sensitivity of estimates to kernel choices and true effect magnitudes. 

\textbf{Scenario D (Pairwise Interactions in G1):} The true model includes pairwise interactions within G1, but only 50\% of possible combinations are active ($\gamma_{jk} \in \{1.5, 0, -1.5\}$). A linear GLIMARK estimates similar contributions for G1 and G2 (19\% vs.18\%), failing to separate interactions from main effects. A hybrid GLIMARK (linear + degree-2 polynomial) better decomposes these effects, assigning 20\% to G1-Poly and 6\% to G2-Linear. This aligns with the polynomial kernel’s feature mapping, which captures two-way interactions without explicit specification.

\textbf{Scenario E (Simplified Interaction Case):} We consider only G1, with five active variables. There are 10 possible two-way interactions, of which 70\% are randomly selected as active. Interaction coefficients \(\gamma_{jk}\) are assigned values from \(\{1.5, 0, -1.5\}\), where nonzero values correspond to active interactions. Two hybrid GLIMARK models are tested: one using a degree-2 polynomial kernel for all interactions and a linear kernel for individual variables, and another replacing the polynomial kernel with an RBF kernel. Both models successfully detect active interactions.

As a  summary of our first aim, multiple kernel GLIMARK model can highlight variables and attribute them to the correct kernels when they are strongly related to the underlying functional forms. But it is also subject to uncertainties and sensitive to mis-specifications.

% use figure* ONLY for 2 col view!!! Other wise we will lose reference due to floating placement.

For our second aim, we used the same variable setting as in the first aim, corresponding to Scenarios F and G (Table~\ref{tbl:tbl1}). This setup simulates complex nonlinear functional forms with both group and individual effects. A grid search for $\lambda$ and $T$ was conducted. For each $\lambda$ value, we simulated 100 training-test dataset pairs and ran a three-kernel GLIMARK with varying $T$ values. The kernels used were linear, degree-2 polynomial, and RBF with a sigma value of 10.

For Binary data, we evaluated model performance based on the accuracy from optimal Youden’s J statistic and the area under the curve (AUC). For count data we used root mean square error (RMSE). For comparison, we also tested a linear SVM, an RBF-kernel SVM, and a random forest model with 100 trees on the same datasets. 

\textbf{Scenario F (Binary Data):} Generally the AUC of GLIMARK increases with $T$ and decreases with $\lambda$. Best AUC is at $T=180$ and $\lambda=10^{-7}$. This suggests binary data in this case benefits from relative weak regularization but more representers. The corresponding accuracy, using the optimal cutoff, is 0.78. In comparison, the accuracy for the linear-SVM model, the RBF-SVM model, and the RF model are 0.65, 0.75, and 0.76, respectively.

\textbf{Scenario G (Count Data):} 
We also simulated Poisson data in Scenario G and found a different pattern. RMSEs are minimized when \( T=6 \); adding more representers beyond this point leads to decreased performance. Similarly, RMSEs increase as \( \lambda \) decreases, suggesting that stronger regularization is essential to prevent overfitting, as opposed to binary data. The lowest RMSE for GLIMARK is 1.48, compared to 1.53 for RF.

\section{Application to Precision Health COVID-19 CXR Data}

 \subsection{Patient Population and Radiomic Features}
Patients included in our study were identified from the Precision Health database and met the following criteria: (1) a confirmed or suspected COVID-19 diagnosis; (2) at least one AP or PA view CXR taken at Michigan Medicine; and (3) an encounter occurring between March 2020 and March 2022. The binary outcome is defined as whether the patient was transferred to any ICU at Michigan Medicine within 30 days of their initial COVID-19 diagnosis. Of the 2,396 patients included, 54\% were male. The racial composition was 69\% White, 20\% African American, and 11\% from other racial groups. The median age was 60 years (IQR: 44–77). A total of 756 patients (31.5\%) required ICU escalation.
Detailed patient demographic and comorbidity information can be found in Supplementary Materials.
For each  patient, only CXRs obtained on the date of first COVID-19 diagnosis were considered, and if multiple were available, only one was included in the analysis.

The images are in Digital Imaging and Communications in Medicine (DICOM) format. On average, each image contains about 2757~$\times$~3195 pixels. Pixel normalization maps the raw values to a range between 0 and 255, and histogram equalization enhances image contrast. Segmentation was accomplished using U-net, a commonly used network architecture \citep{ronneberger2015u}. With ground truth CXR images and corresponding masks, U-net can achieve good segmentation. A segmentation mask is a binary image where each pixel corresponds to the same location in the original image and is labeled to indicate whether it belongs to the lung region or the background, both considered as parts of the ROI. These masks are often generated by clinicians. We used two public chest X-ray datasets with lung masks: the Montgomery County (MC) dataset and a 55-image subset of the Indiana University dataset with manual annotations \citep{candemir2013lung, xue2023cross}. Web Figure~3 illustrates the image processing pipeline.

We use PyRadiomics the package \citep{van2017computational} to extract features. The workflow includes a filtering step followed by a feature extraction step. First we apply one of the eleven filters to each image to enhance it. Then according to a fixed formula, we extract features belonging to one of the seven classes for either lung segmentation or background segmentation. Web Table 3 lists all filters and classes. For example the First-Order Statistics class has nineteen features and Gray Level Run Length Matrix class has sixteen features. With all filters and classes, each image can yield 1032 features per segmentation and 2064 in total. Web Figures 1–2 show exploratory plots summarizing univariate associations with ICU escalation. For the GLIMARK model, we present the hierarchical structure as a three-level framework. Level 1 contains all features based on segmentation. Level 2 contains all features based on classes. Level 3 contains individual features (Figure~{\ref{fig:concept}}).

\subsection{Model Levels and Kernel Choices}

To determine the hierarchical structure of our model, we conduct a stepwise exploratory analysis across nested model stages, as illustrated in Figure~\ref{fig:concept}. First, we fix \( \lambda = 0.001 \) and \( T = 400 \), as was done in the simulation study. Next, we run a one-level GLIMARK model at the segmentation level, using a degree-2 polynomial kernel with an offset of one, and RBF kernels with a broad range of \( \sigma \) values. We then select the optimal \( \sigma \) for the RBF kernel in combination with the polynomial kernel and finalize it as Model 1. Building upon it, we incorporate feature group-level data and propose new RBF kernels with varying \( \sigma \) values, again in combination with the polynomial kernel then finalize it with the best RBF kernel (Model 2). Finally, we expand the top group-level features in Model 2 and include individual features from these groups as level-three data, with the same polynomial kernel and one RBF kernel with smaller \( \sigma \) (Model 3). The rationale is to control model complexity in terms of $P$ and $Q$ while enabling the model to explore patterns across different levels. This feature exploration step is applied to the entire dataset. After finalizing the hierarchical structure, we perform 5-fold cross-validation on all three models to evaluate and report the average performance.

\subsection{Results and Conclusions}
\subsubsection{Exploratory Analysis Results}
The detailed results for the nested exploratory analysis are provided in  Web Appendix B. Briefly, Model 1 demonstrates significant contributions from RBF and polynomial kernels in both segments. By Model 2, feature class terms become dominant, significantly reducing contributions from segmentation levels. Four feature classes are included for Model 3 on individual feature level, which is considered the final model and cross-validated.

\subsubsection{Model Comparison and Feature Importance}

We compare the 5-fold cross-validation fine-tuning performance of the GLIMARK model against RFs and the logistic regressions with $\ell_{1}$ and $\ell_{2}$ regularization. For RFs and logistic regressions all features were used. 

\begin{figure}
    \centering
    \includegraphics[height=0.9\textheight, keepaspectratio]{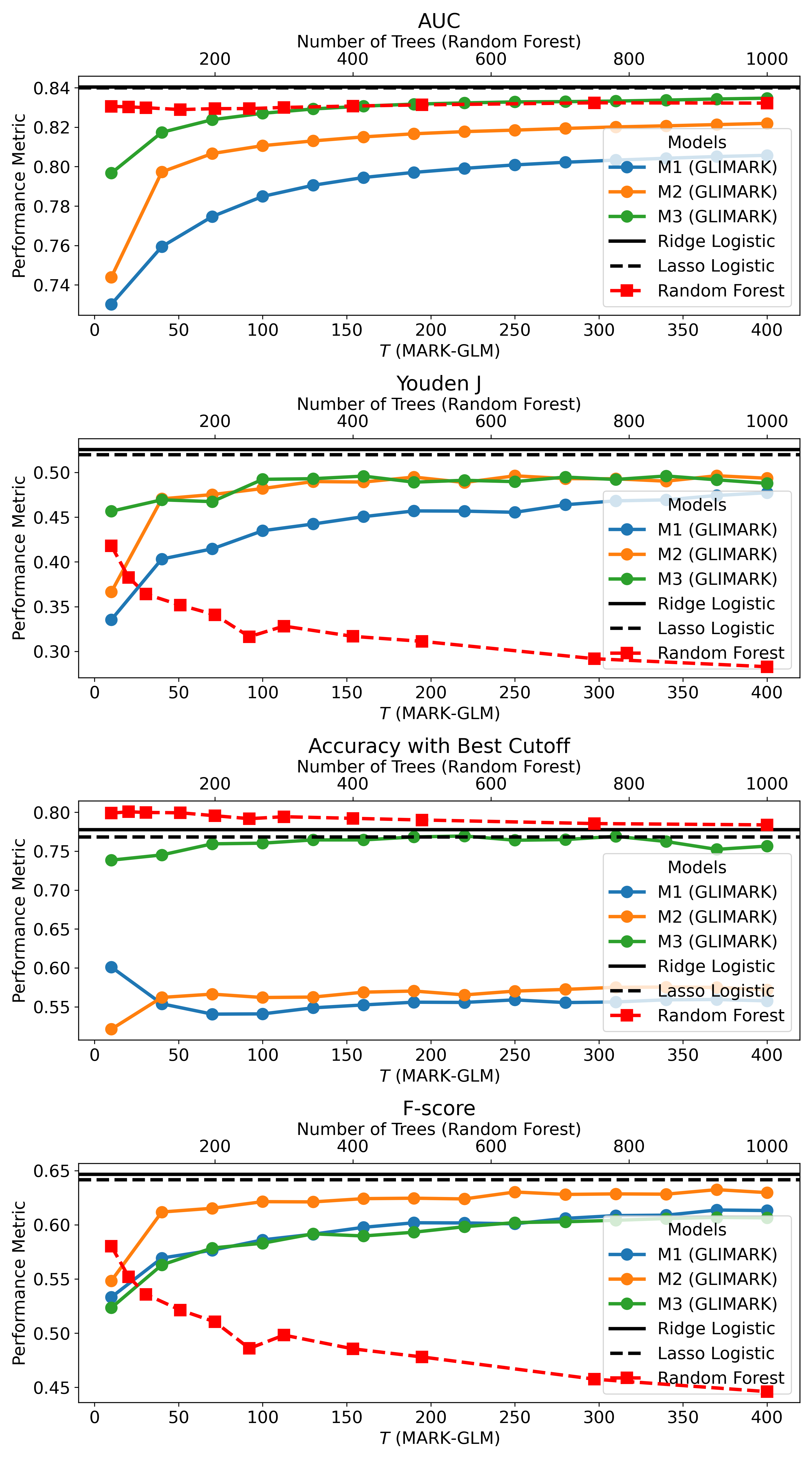}
    \caption{Performance comparison. For Random Forest, each point represents the average cross-validation performance for a fixed \( n_{\text{estimators}} \). For GLIMARK, since performance depends on both \( \lambda \) and \( T \), each point represents the highest average cross-validation performance obtained across different \( \lambda \) values for a given \( T \).
}
    \label{fig_4grid}
\end{figure}

Figure~{\ref{fig_4grid}} shows that all three GLIMARK models have increased performance as $T$ increases. Model 3 begins to exhibit higher AUC than RFs when GLIMARK reaches approximately 200 representers. The highest AUC observed for Model 3 is around 0.83. GLIMARK models also have generally higher Youden's J index and F-scores than RFs.  It is also noteworthy that GLIMARK models with fewer than 400 terms perform on par with fine-tuned logistic regressions using all 2046 predictors across all metrics. This suggests that GLIMARK effectively balances parsimony and predictive power.

\begin{figure}
\centerline{%
\includegraphics[width=\columnwidth]{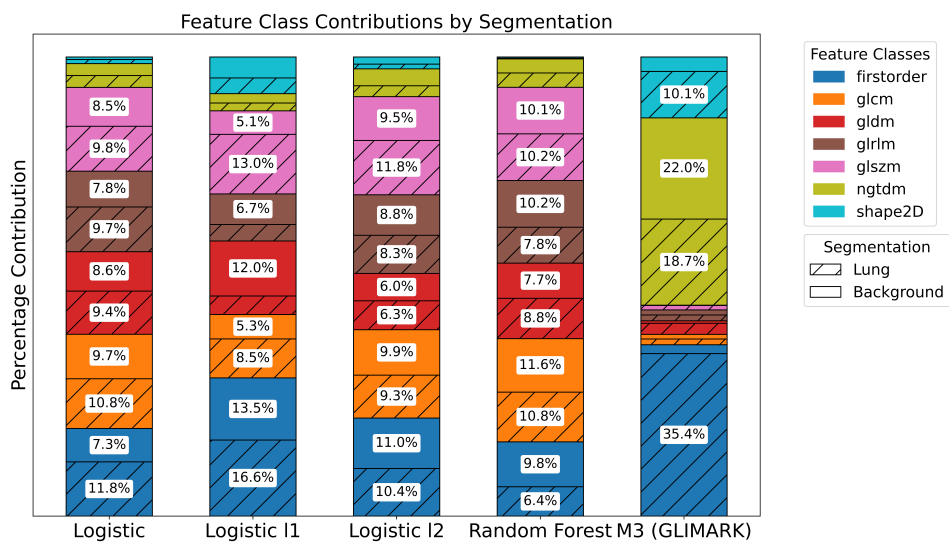}}
%% to include a figure, or to leave a blank space
\caption{Feature importance. For logistic models, contribution percentages are computed as the sum of absolute coefficient values within each class. For Random Forest, importance is based on the sum of Gini importance within each class. All models, except the basic logistic regression, correspond to their best fine-tuned configurations. For the GLIMARK model, Level 3 estimates are merged back to their corresponding Level 2 feature classes and we omit the small contributions from Level 1, as Level 1 does not provide feature class estimates. }
\label{fig_stackbar}
\end{figure}

Figure~{\ref{fig_stackbar}} shows the feature importance across different models. The behavior of the Model 3 is close to that of lasso-regularized logistic regression in that it highlights fewer more important feature classes. GLIMARK further focuses on  three dominant classes: First-Order, NGTDM, and Shape2D.
These findings are consistent with existing literature. First-Order radiomic features have been shown to predict ICU admission and intubation in COVID-19 patients \citep{varghese2021predicting}, aligning with findings that increased pixel intensity in lung regions corresponds to COVID-19-related lung involvement \citep{al2023covid}. Shape2D\_L features measure the shape of lung area, and it is shown that severe COVID-19 patients have deformities to the surfaces of the lungs \citep{HIREMATH2024108643}. Previous studies have not extensively reported NGTDM features in COVID-19 prognosis, but our analysis found them to be significant predictors as well.

% Must use * if the figure needs to span 2 cols!! For referee option cannot use *

\subsubsection{Representers}
\begin{figure} % Use [H] for exact placement
\centerline{%
\includegraphics[height=0.9\textheight, keepaspectratio]{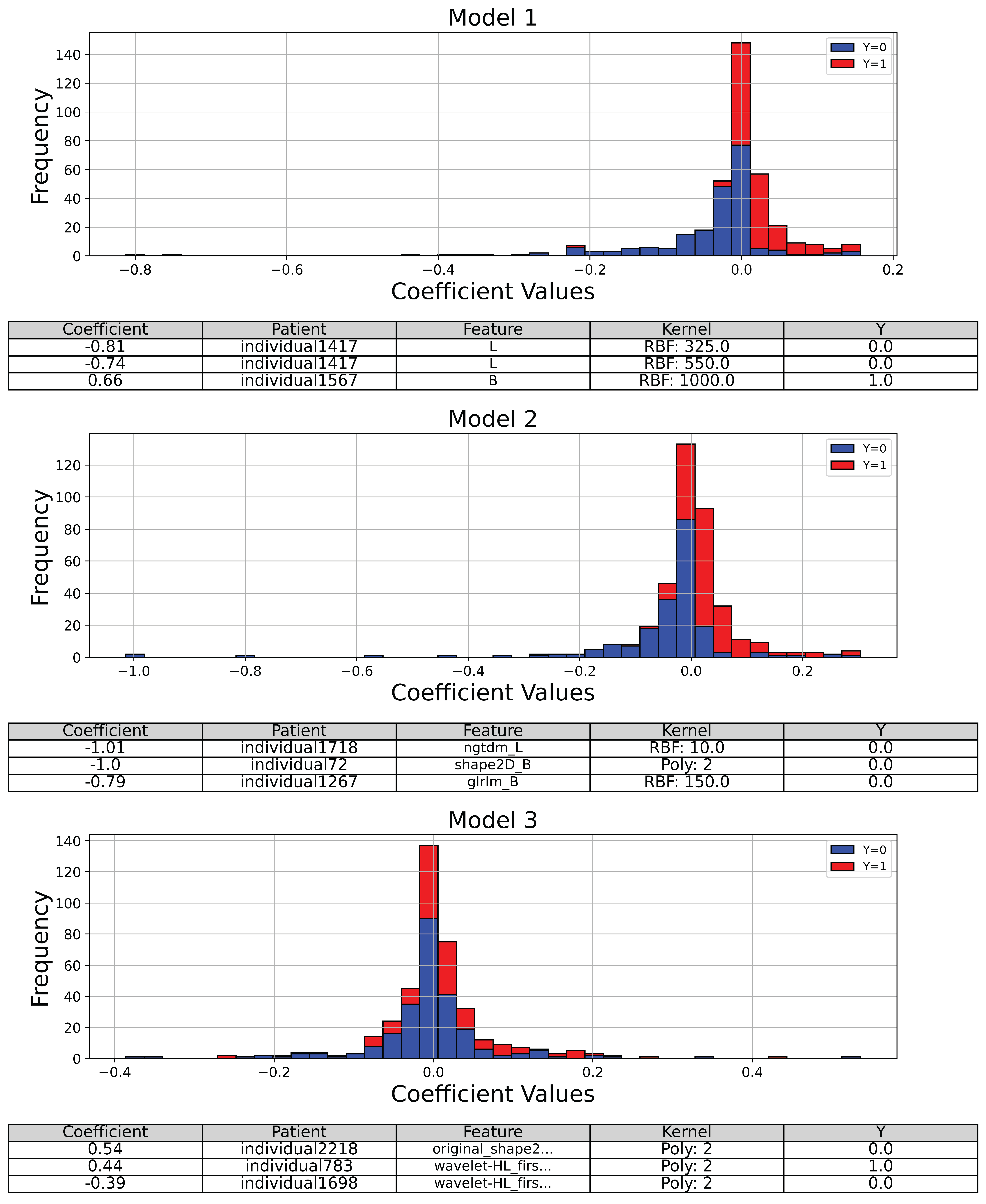}}
  \caption{Coefficient contribution and top representers. For each model only the three most influential points are displayed. }
  \label{fig_representer}
\end{figure}

We consider Model 3 the best model due to its overall performance. However, in this section, we revisit all three models to highlight their distinct and complementary findings. The GLIMARK model's interpretability stems from its similarity to existing data points, i.e., positive $\alpha$ coefficients are expected to correlate with ICU escalation cases. To assess this, we compared the signs of the $\boldsymbol{\hat{\alpha}}$ coefficients with the true labels of the training points for three models in Figure~{\ref{fig_representer}}. With 400 representers, Model 1 had the greatest agreement rate (93\%), followed by Model 2 (87\%). Model 3 only had 60\% agreement rate. One explanation is that both cosine similarity and Euclidean similarity provide limited interpretability when comparing two points based on a single attribute.

 Another interesting finding was that the top representers varied depending on the scale of the data. Figure~{\ref{fig_representer}} shows that in Model 1, patient 1417 is the most influential across both segments. However, when features are further categorized by classes in Model 2, patient 1417 is no longer among the top ten representers; instead, patient 1718 with NGTDM\_L emerges as the most influential. Similarly, transitioning from Model 2 to Model 3 alters the most influential patients, emphasizing the impact of individual feature selection.

 Since kernel selection and feature aggregation methods are not unique, our analysis of CXR images is exploratory rather than definitive, focusing on potential relationships between image features and ICU outcomes. To support this, we present CXR images for the top representers in each model in Web Figure 4-6, aiming to guide future research by professional radiologists and other interested researchers in interpreting visual features linked to ICU escalation.
\subsection{Conclusion}
GLIMARK achieves performance comparable to both RFs and logistic regressions with approximately 100 representers, significantly reducing model complexity. It identifies First-Order, NGTDM, and Shape2D as the most important feature classes associated with the outcome, whereas other models suggest a more uniform distribution of feature importance.  These findings align with radiology literature. Furthermore, GLIMARK enhances interpretability by highlighting the most representative patients associated with each feature.

\section{Discussion}

One strength of hierarchical column generation methods like GLIMARK is its flexibility in exploring features on different scales. Models using high-level groups of features explore potential interactions between features, whereas low-level models implicitly select features. It extends the complexity of the model with a single kernel using all features. Unlike KPCA, it does not map features, thus retaining a certain degree of interpretability, especially when we use handcrafted features.

There are a few open research directions. First, although our strategy for the CXR data is a multi-view learning approach, it didn't fully utilize the potential of GLIMARK for heterogeneous data. Segmentation based on organs, though common, does not further explore localized patterns on smaller scales. Recently, with the popularity of visual transformers (ViTs) \citep{dosovitskiy2020image}, researchers have proposed numerous methods to focus on smaller batches of images. High-quality CXR images in our dataset enable transformer models to generate attention maps focused on more detailed parts such as airway, heart or diaphragm. With ViTs, identifying ROIs for LAT images is also possible. Thus, a more systematic multi-view strategy can be structured as a grid, where one axis represents CXR perspectives with their attention ROIs (e.g., PA/AP vs. LAT), another axis represents radiomic filters, and a third axis represents radiomic feature classes.

Secondly, our architecture for aggregating features and applying kernels is not exhaustive. Domain-specific kernels have been extensively studied \citep{shervashidze2011weisfeiler, lodhi2002text}. For CXR images, however, relative kernels are limited and warrant more research. Moreover, handcrafted features are not the only way to extract features. For instance, many deep learning features such as ResNet-50 have also been largely used. Future work could focus on incorporating both radiomics and DNN features.

\section*{Acknowledgments}
This study was supported by the National Cancer Institute under grant numbers R01CA269398 and R01CA249096. AI-assisted tools (e.g., ChatGPT) were used to support partial code development; These tools were not used for writing or editing the manuscript text. We thank Dr. Stephen Salerno and Dr. Yuming Sun for their guidance and for sharing code used to prepare the Michigan DataDirect dataset.

\section*{Data Availability}
The data that support the findings of this study are available from Michigan DataDirect. Restrictions apply to the availability of these data, which were used under license for this study. Data are available upon request from Michigan DataDirect (PHDataHelp@umich.edu) with the necessary permissions.

%-------------------------------------------
% References
%-------------------------------------------

\bibliographystyle{plainnat}
\bibliography{ref} % This assumes your reference file is named ref.bib

\section*{Supplementary Materials}
Additional tables, figures, and technical details are provided in the accompanying Supplementary Materials.

%-------------------------------------------
% Appendix
%-------------------------------------------
% Activate the appendix in the doc
% from here on sections are numerated with capital letters 
%\appendix

% % Change equation numbering format to be sequential within sections in the appendix
% \renewcommand\theequation{\Alph{section}\arabic{equation}} % Redefine equation numbering format
% \counterwithin*{equation}{section} % Number equations within sections
% \renewcommand\thefigure{\Alph{section}\arabic{figure}} % Redefine equation numbering format
% \counterwithin*{figure}{section} % Number equations within sections
% \renewcommand\thetable{\Alph{section}\arabic{table}} % Redefine equation numbering format
% \counterwithin*{table}{section} % Number equations within sections

\end{document}